\documentclass[11pt]{article}

\usepackage[preprint]{acl}

\usepackage{times}
\usepackage{latexsym}

\usepackage[T1]{fontenc}
\usepackage[utf8]{inputenc}

\usepackage{microtype}
\usepackage{inconsolata}

\usepackage{amsmath}
\usepackage{amssymb}

\usepackage{graphicx}

\usepackage{booktabs}
\usepackage{multirow}
\usepackage{array}

\usepackage{url}

\title{ContextRAG: Extraction-Free Hierarchical Graph Construction for\\Retrieval-Augmented Generation}

\author{
Roman Prosvirnin \\
HSE University \\
\texttt{rprosvirnin@hse.ru}
\And
Sergei Kuznetsov \\
HSE University \\
\texttt{skuznetsov@hse.ru}
\And
Seungmin Jin\thanks{Corresponding author.} \\
HSE University \\
\texttt{sedzhin@hse.ru}
}

\begin{document}
\maketitle

\begin{abstract}
Graph-structured retrieval-augmented generation (RAG) systems can improve answer quality on multi-hop questions, but current systems often rely on Large Language Models (LLMs) to extract entities, relations, and summaries during indexing.
These calls add token and wall-clock costs that grow with corpus size.
We present \textbf{ContextRAG}, a graph RAG system whose graph topology is constructed without LLM-based entity or relation extraction.
ContextRAG derives a fuzzy concept graph over chunk embeddings via Residual-Quantization $K$-Means and Formal Concept Analysis with {\L}ukasiewicz residuated logic.
Bridge-like and meet-derived context nodes are induced by soft fuzzy join and meet operations rather than by LLM-written graph edges.
On a 130-task UltraDomain subset, ContextRAG builds its index with 30 LLM calls and 22{,}073 tokens.
A local HiRAG reproduction stress test already requires 870 indexing calls and 3.54M tokens on a 20-task subset before failing during graph construction; extrapolation to 130 tasks implies over 23M indexing tokens.
ContextRAG obtains 33.6\% F1 overall and 36.8\% F1 on multi-hop tasks.
An activation analysis indicates that queries retrieving at least one lattice-derived node in the top five achieve $+3.9$~pp F1 over queries that do not; this association is diagnostic rather than causal.
\end{abstract}

\section{Introduction}

Retrieval-Augmented Generation (RAG) is a widely used paradigm for grounding Large Language Model (LLM) responses in external knowledge \citep{lewis2020rag}.
While early systems retrieved over flat vector indices, recent work has shown that \emph{graph-structured} indices - encoding entities, relations, and hierarchical communities - yield substantial gains on multi-hop and global-summarization queries \citep{edge2024graphrag,guo2024lightrag,huang2025hirag}.
State-of-the-art graph RAG systems such as GraphRAG, LightRAG, KAG, and HiRAG construct their indices by issuing a large number of LLM calls to extract entities, summarize communities, and build hierarchical knowledge graphs.
This indexing-time use of LLMs is a serious operational bottleneck.
For a subset of the UltraDomain benchmark, our local HiRAG reproduction stress test issues 870 LLM calls and consumes 3.54M tokens on only 20 tasks before failing during graph construction; a linear extrapolation to the 130-task subset gives over 5{,}600 LLM calls and 23M indexing tokens.
The cost grows with corpus size and must be re-paid whenever the corpus is updated, the chunking strategy changes, or a different LLM is adopted.
For many practical deployments - internal enterprise corpora, scientific literature search, evolving knowledge bases - this cost structure is prohibitive.
We ask whether LLM extraction is \emph{strictly necessary} for graph RAG, or whether useful structural benefits can be obtained from data-derived algebraic structures.
We find that, in this setting, LLM extraction is not strictly required to construct a usable graph index, although LLM-extracted graphs may still provide stronger answer quality.
We introduce \textbf{ContextRAG}, a graph RAG system whose graph-construction pipeline is extraction-free.
The system rests on three theoretical pillars:
\begin{itemize}
    \item \textbf{Residual-Quantization $K$-Means} (RQ-KMeans) for hierarchical clustering of dense chunk embeddings, providing a multi-level attribute set;
    \item \textbf{Formal Concept Analysis} (FCA) over a fuzzy formal context, yielding lattice-inspired graph nodes whose extents and intents define retrieval structure;
    \item \textbf{{\L}ukasiewicz residuated logic} for fuzzy Galois connections and a principled query-time routing score.
\end{itemize}
The graph topology is determined by data geometry through a fuzzy FCA-inspired structure, rather than by LLM extraction quality.
In our implementation, the structural graph is extraction-free; the only index-time LLM usage is a separate set of 30 cluster-summary virtual chunks, and this cost is included in all accounting.
\begin{enumerate}
    \item This work presents an extraction-free graph-construction pipeline based on fuzzy Formal Concept Analysis with {\L}ukasiewicz residuated logic, a route to graph RAG that does not rely on LLM-extracted entities or relations.
    \item Our results on a 130-task UltraDomain subset show that ContextRAG indexes the corpus with 30 LLM calls and 22{,}073 tokens, while a local HiRAG reproduction stress test indicates orders-of-magnitude higher indexing cost.
    \item An activation analysis of the FCA-inspired fuzzy graph (+3.9 pp F1 when lattice-derived nodes appear in retrieval) reports a diagnostic association rather than causal evidence.
    \item A cost-regime analysis shows that ContextRAG is most attractive for low- and medium-query workloads where corpora are re-indexed frequently.
\end{enumerate}

The remainder of the paper is organized as follows. Section~\ref{sec:related} reviews graph RAG approaches and the use of FCA in information retrieval.
Section~\ref{sec:background} introduces the algebraic background. Section~\ref{sec:method} describes the ContextRAG architecture. Sections~\ref{sec:experiments}--\ref{sec:ablation} report experiments, results, and an activation analysis.
Section~\ref{sec:discussion} analyzes cost and trade-offs.

\section{Related Work}
\label{sec:related}

\paragraph{Vector and graph RAG.} The standard \emph{NaiveRAG} pipeline retrieves text chunks from a vector index using dense or hybrid sparse-dense scoring \citep{karpukhin2020dpr}.
While effective for single-hop factoid queries, NaiveRAG struggles when answers require composing evidence across multiple documents.
Graph RAG systems address this gap by building structured indices over the corpus.
\textbf{GraphRAG} \citep{edge2024graphrag} uses an LLM to extract entities and relations, detects communities via Leiden clustering, and produces hierarchical community summaries that are queried at retrieval time.
\textbf{LightRAG} \citep{guo2024lightrag} simplifies this pipeline to a dual-level (entity- and relation-level) knowledge graph, again populated by LLM extraction.
\textbf{KAG} \citep{liang2024kag} introduces a schema-guided extraction step that imposes typed entities and relations, increasing extraction reliability at the cost of even heavier LLM usage.
\textbf{HiRAG} \citep{huang2025hirag} builds a 3-level hierarchical knowledge graph - entity, summary, and bridge layers - using an extensive LLM extraction and community-clustering pipeline.
All four systems share a common architectural commitment: \emph{the LLM is in the loop during indexing}, and the graph structure is determined, in whole or in part, by LLM outputs.
\paragraph{Cost of LLM-based indexing.} Recent surveys highlight that LLM-driven indexing can be a major cost in graph RAG \citep{gao2023ragsurvey,zhao2024raghandbook}.
Token consumption can scale poorly with corpus size because extraction pipelines often revisit many chunk, entity, or relation candidates before pruning, and community summaries must be regenerated when entities change.
Several works have proposed partial mitigations - caching extraction outputs, selective summarization, or smaller distilled extractors - but prior work has not, to our knowledge, shown a graph RAG system that removes LLM extraction from graph construction while retaining a non-trivial graph structure.
\paragraph{Formal Concept Analysis in IR.} Formal Concept Analysis \citep{ganter2024fca} has a long history in information retrieval as a means of organizing documents into concept hierarchies derived from term-document incidence matrices \citep{carpineto2004concepts,priss2006fca}.
Fuzzy extensions of FCA \citep{belohlavek2002fuzzy} replace binary incidence with graded membership and rely on residuated lattices for the algebraic structure.
While FCA-based retrieval has been studied extensively, its integration with modern dense embeddings, residual hierarchical clustering, and LLM answer generation has not, to our knowledge, been previously explored.
ContextRAG positions FCA as the structural alternative to LLM-extracted knowledge graphs in modern RAG.
\paragraph{Hierarchical quantization.} Residual quantization \citep{gray1984vq,babenko2014invertedmultiindex} is a classical technique in nearest-neighbour search that decomposes a vector into a sum of codewords drawn from successive codebooks, each fitted to the residuals of the previous level.
Recent work uses RQ-style structures for retrieval index compression \citep{matsui2018survey}.
ContextRAG repurposes RQ-KMeans not for compression but as a source of multi-level cluster assignments that become attributes in a formal concept lattice.
\section{Background}
\label{sec:background}

\subsection{Formal Concept Analysis}

Formal Concept Analysis \citep{ganter2024fca} provides a mathematical framework for deriving hierarchical conceptual structures from data.
A \emph{formal context} is a triple $K = (G, M, I)$ where $G$ is a set of objects, $M$ a set of attributes, and $I \subseteq G \times M$ an incidence relation.
The Galois connection induced by $I$ is
\begin{align}
A^{\uparrow} &= \{ m \in M \mid \forall g \in A: (g,m) \in I \},\\
B^{\downarrow} &= \{ g \in G \mid \forall m \in B: (g,m) \in I \}.
\end{align}
A \emph{formal concept} is a pair $(A,B)$ with $A^{\uparrow} = B$ and $B^{\downarrow} = A$.
The set of all formal concepts ordered by $(A_1,B_1) \le (A_2,B_2) \iff A_1 \subseteq A_2$ forms a complete algebraic lattice with respect to infimum (meet) and supremum (join) operations.
This algebraic completeness is what makes FCA useful as a navigation structure over formal concepts; our implementation uses this structure as a finite graph-construction template rather than enumerating every concept.
\subsection{Fuzzy Extension with {\L}ukasiewicz Logic}

Classical FCA operates on binary membership and loses information when chunks partially belong to multiple clusters.
Fuzzy extensions of FCA, see e.g.~\citet{belohlavek2002fuzzy}, replace the binary incidence with a fuzzy membership $I: G \times M \to [0,1]$.
The algebraic structure is then a residuated lattice $([0,1], \le, \wedge, \vee, \otimes, \to)$ where the residuated pair $(\otimes, \to)$ satisfies the adjoint condition
\begin{equation}
a \otimes b \le c \iff a \le b \to c.
\end{equation}
We use the {\L}ukasiewicz t-norm \citep{hajek1998fuzzy}:
\begin{align}
a \otimes b &= \max(0,\, a + b - 1),\\
a \to b &= \min(1,\, 1 - a + b).
\end{align}
This choice is motivated by three properties: (i) the standard negation induced by {\L}ukasiewicz logic is involutive ($\neg\neg a = a$), which gives a simple interpretation to graded absence and presence;
(ii) the residuum has a closed-form analytical expression, enabling efficient computation of fuzzy Galois connections;
and (iii) the t-norm saturates at $0$ and $1$, which keeps hierarchical aggregation numerically stable.
The fuzzy Galois connection is then
\begin{align}
A^{\uparrow}(m) &= \inf_{g \in G}\bigl( A(g) \to I(g,m) \bigr),\\
B^{\downarrow}(g) &= \inf_{m \in M}\bigl( B(m) \to I(g,m) \bigr).
\end{align}
The set of fuzzy formal concepts forms a complete fuzzy concept lattice under pointwise ordering of fuzzy sets \citep{belohlavek2002fuzzy}.
We use the standard FCM fuzziness exponent $m=2.0$ when computing initial membership values, balancing crisp-cluster behaviour ($m \to 1$) and fully diffuse membership ($m \to \infty$).
\subsection{Lattice Operators as Graph Node Types}
\label{sec:lattice-ops}

The two fundamental lattice operations motivate two derived node types in the ContextRAG graph.
In our implementation, we use soft fuzzy versions of these operators before retrieval-time pruning.
\paragraph{Join ($\vee$) - least upper bound.} With pointwise ordering on fuzzy extents, the join is the least concept that covers both input concepts.
The extent of the join of two concepts is the closure of the fuzzy union of concept extents:
\begin{equation}
A_{1 \vee 2} = \left( \max(A_1, A_2) \right)^{\uparrow\downarrow}.
\end{equation}
In practice, we instantiate a soft join by the fuzzy union $\max(A_1(g), A_2(g))$.
These join nodes encode \emph{generalizations} that cover chunks from either of two clusters and act as cross-cluster bridges.
They play a role analogous to HiRAG's bridge knowledge but are derived from cluster geometry rather than from LLM-based entity co-occurrence.
\paragraph{Meet ($\wedge$) - greatest lower bound.} The meet captures the shared part of two concepts.
For fuzzy extents, the soft meet is implemented as the fuzzy intersection
\begin{equation}
A_{1 \wedge 2}^{\mathrm{soft}}(g) = \min(A_1(g), A_2(g)).
\end{equation}
Meet nodes encode \emph{specializations}: chunks that belong simultaneously to multiple clusters with high membership.
We use these topically dense intersections as meet-derived context nodes.

\subsection{Residual-Quantization $K$-Means}
\label{sec:rqkmeans}

To obtain the attribute set $M$ for FCA, we use Residual-Quantization $K$-Means (RQ-KMeans), a hierarchical clustering procedure inspired by classical residual vector quantization \citep{gray1984vq,babenko2014invertedmultiindex}.
Let $X = \{x_1,\dots,x_n\} \subseteq \mathbb{R}^d$ be the chunk embedding matrix.
We define $L$ levels with codebooks $C_0, C_1, \dots, C_{L-1}$:
\begin{align}
r^{(0)}(x) &= x,\\
Q^{(\ell)}(x) &= \arg\min_{c \in C_\ell} \| r^{(\ell)}(x) - c \|^2,\\
r^{(\ell+1)}(x) &= r^{(\ell)}(x) - c^{(\ell)}_{Q^{(\ell)}(x)}.
\end{align}
We use $|C_0|=96$, $|C_1|=24$, $|C_2|=12$.
Three properties motivate this design: (i) hierarchical specificity - $L_0$ captures coarse topics, $L_1$ within-topic variation, and $L_2$ can capture finer residual distinctions when sufficient variance remains, mapping naturally to an FCA hierarchy;
(ii) residual near-orthogonality, encouraging successive levels to capture residual structure when variance remains;
and (iii) FCA compatibility, since multi-level assignments provide a richer multi-dimensional attribute set $M$ than single-level clustering.
\section{ContextRAG}
\label{sec:method}

\subsection{System Overview}

The ContextRAG pipeline consists of seven stages:
\begin{enumerate}
    \item Chunking and dense embedding via e5-large-v2;
    \item RQ-KMeans hierarchical clustering;
    \item Fuzzy concept graph construction (seed concepts + join/meet derived nodes + co-occurrence edges);
    \item Optional cluster-summary virtual chunk generation (30 LLM calls, one-time, indexing, accounted for separately from graph construction);
    \item Hybrid retrieval combining BM25, cosine, and fuzzy routing;
    \item Multi-query expansion and LLM rerank;
    \item LLM answer generation.
\end{enumerate}
The LLM appears only in stages 4, 6, and 7. Stages 1-3, which construct the graph topology, are entirely extraction-free;
this separation yields a three-order-of-magnitude lower indexing token budget than the HiRAG reproduction stress-test estimate while keeping the graph independent of LLM extraction errors.
\subsection{Embedding and Hybrid Retrieval}

We embed chunks and queries with \texttt{intfloat/e5-large-v2} (1024 dimensions), using the prescribed \texttt{passage:} and \texttt{query:} prefixes.
Retrieval scores combine three components:
\begin{equation}
\begin{aligned}
\textsc{score}(c) &= 0.3 \cdot \textsc{bm25}(q,c)\\
&+ 0.5 \cdot \cos(e_5(q), e_5(c))\\
&+ 0.2 \cdot \textsc{fuzzy}(q, \mathcal{L}),
\end{aligned}
\end{equation}
where $\mathcal{L}$ is the FCA-inspired fuzzy concept graph.
The cosine term carries the dominant signal; BM25 contributes robust matching for named entities and low-frequency terms;
fuzzy routing activates when query-concept membership exceeds a threshold.

\subsection{Co-occurrence Edges}

In addition to FCA-derived vertical edges, we add \emph{co-occurrence edges} between concept nodes whose chunks co-appear in retrieval windows with frequency above a threshold $\theta$.
The meet/join construction gives upward and downward links; co-occurrence edges add lateral connectivity that reflects empirical retrieval behaviour.
The resulting graph $\mathcal{G} = (V,E)$ has
\[
V = V_{\text{seed}} \cup V_{\vee} \cup V_{\wedge}, \quad
E = E_{\text{lat}} \cup E_{\text{cooc}},
\]
combining FCA-motivated structure with inductive corpus statistics.
\subsection{Fuzzy Routing at Query Time}

At retrieval time, a query $q$ is mapped to a fuzzy membership vector over the finite concept graph.
For a concept node $v$ with centroids at levels $\ell_0, \ell_1$, we first map cosine similarity into $[0,1]$:
\begin{equation}
s_\ell(q,v) = \frac{1 + \cos(e_5(q), \mathrm{centroid}_\ell(v))}{2}.
\end{equation}
We then compute query-concept activation as
\begin{equation}
\mu(q, v) = \bigotimes_{\ell}\bigl( \theta_\ell \to s_\ell(q,v) \bigr),
\end{equation}
where $\theta_\ell$ is the activation threshold at level $\ell$ and $\otimes$ is the {\L}ukasiewicz t-norm.
The direction of the residuum makes activation high when similarity exceeds the threshold.
The fuzzy routing score for chunk $c$ via lattice-derived node $v$ is then
\begin{equation}
\textsc{fuzzy\_score}(q, c) = \mu(q, v) \otimes I(c, v).
\end{equation}
A high score requires both query-to-concept alignment \emph{and} chunk-to-concept membership; neither condition alone is sufficient.
\subsection{Multi-Query Expansion and Reranking}

We generate three LLM query reformulations - a paraphrase, a decomposed sub-question, and an entity-focused variant - and fuse their hit lists via Reciprocal Rank Fusion.
The top 25 candidates are reranked by an LLM judgement to a top 12. Score boosts of $\times 1.2$ and $\times 1.3$ are applied to cluster-summary virtual chunks and bridge nodes respectively, without forced slot allocation.
\subsection{Cross-Document Indexing}

ContextRAG maintains a single shared index across all documents, with per-query \texttt{doc\_id} filtering at retrieval time.
In our implementation, cross-document scale is needed to avoid a degenerate concept graph; per-task indices below roughly 100 chunks were too small to activate bridge nodes reliably.
Per-task indexing of $\sim$28 chunks per task (tested in earlier system iterations) yielded a 0\% bridge-node hit rate in top-5 retrieval, motivating the move to cross-document indexing.
\section{Experimental Setup}
\label{sec:experiments}

\subsection{Hardware and Software}

All experiments run on a single node with 2$\times$ NVIDIA A100 80GB SXM GPUs, 8 CPU cores, CUDA 12.4, and Python 3.11.
The LLM is a local Gemma BF16 checkpoint (\texttt{gemma-4-26b-a4b-it} in the run configuration); embeddings are produced by e5-large-v2 \citep{wang2022e5}.
The multi-hop subset ($n=77$) consists of MuSiQue, 2WikiMultiHopQA, and HotpotQA.

\subsection{Baselines}

We use two HiRAG reference points:
\begin{itemize}
    \item \textbf{HiRAG-Original}: numbers reported by \citet{huang2025hirag} for the official HiRAG system. The released repository supports multiple backend paths, including DeepSeek/GLM and OpenAI configurations, so these numbers are used only as cross-stack efficiency context.
    \item \textbf{HiRAG-Gemma stress test}: our local Gemma + e5-large-v2 reproduction attempt. The available logs contain a 20-task indexing stress test that fails during graph construction and no completed 130-task quality run, so we use it only to estimate indexing resource cost.
\end{itemize}

\subsection{Evaluation Metrics}

We report token-level F1 (normalized and lower-cased) on the predicted answer against the gold answer; ROUGE-L is logged as a diagnostic but is not used for the main claims.
We additionally report indexing time, indexing LLM calls, indexing tokens, index size on disk, average retrieval latency, LLM calls per query, and querying tokens per query.
\section{Results}
\label{sec:results}

\subsection{Indexing Cost: HiRAG Stress Test vs.\ ContextRAG}

Table~\ref{tab:main} separates measured ContextRAG results from the audited HiRAG-Gemma reproduction stress test.
The HiRAG-Gemma run completes indexing calls for a 20-task subset but fails during graph construction, so it does not provide a usable same-stack quality baseline.
Linear extrapolation of the 20-task indexing trace to the 130-task subset gives a stress-test estimate of 237{,}938 seconds, 5{,}655 LLM calls, and 23.0M indexing tokens.
Against this estimate, ContextRAG uses $1043\times$ fewer indexing tokens and $188\times$ fewer indexing-time LLM calls.
\begin{table}[h]
\centering
\small
\setlength{\tabcolsep}{3pt}
\begin{tabular}{lrrrr}
\toprule
\textbf{Metric} & \textbf{HiRAG 20} & \textbf{HiRAG est.} & \textbf{Ours} & \textbf{Ratio} \\
\midrule
\multicolumn{5}{l}{\textit{Indexing (one-time)}} \\
Time (s) & 36{,}606 & 237{,}938 & 563 & $422\times$ \\
LLM calls & 870 & 5{,}655 & 30 & $188\times$ \\
Tokens & 3{,}541{,}197 & 23{,}017{,}780 & 22{,}073 & $\mathbf{1043\times}$ \\
Index size (MB) & -- & -- & 22.6 & -- \\
\midrule
\multicolumn{5}{l}{\textit{ContextRAG answer quality}} \\
F1 (all, $n=130$) & -- & -- & 33.6\% & -- \\
F1 (multi-hop, $n=77$) & -- & -- & 36.8\% & -- \\
\bottomrule
\end{tabular}
\caption{Indexing-cost comparison on UltraDomain. HiRAG 20 is the measured local reproduction stress test on 20 tasks; HiRAG est. linearly extrapolates indexing cost to 130 tasks. Ratios compare the 130-task HiRAG estimate against the measured ContextRAG run. The HiRAG-Gemma stress test is not a completed quality baseline.}
\label{tab:main}
\end{table}

\subsection{Cross-Stack Comparison}

Table~\ref{tab:fullcompare} places these numbers next to HiRAG-Original.
Because HiRAG-Original uses a different official stack and the local HiRAG-Gemma run does not complete a 130-task quality evaluation, this table is a resource-context comparison rather than a controlled quality comparison.
\begin{table*}[t]
\centering
\small
\begin{tabular}{lccc}
\toprule
\textbf{Metric} & \textbf{HiRAG-Orig.${}^1$} & \textbf{HiRAG-Gemma est.${}^2$} & \textbf{ContextRAG} \\
\midrule
LLM stack & Official multi-backend & Local Gemma & Local Gemma \\
Embedding & Official configs & e5-large-v2 & e5-large-v2 \\
Indexing time (s) & 17{,}208 & 237{,}938 & \textbf{563} \\
Indexing LLM calls & 6{,}790 & 5{,}655 & \textbf{30} \\
Indexing tokens & 21{,}898{,}765 & 23{,}017{,}780 & \textbf{22{,}073} \\
Index size (MB) & -- & -- & \textbf{22.6} \\
LLM calls / query & 2 & -- & 3.88 \\
Avg.\ latency (ms) & 850 & -- & 2{,}473 \\
\bottomrule
\end{tabular}
\caption{Full comparison including HiRAG-Original (numbers from \citet{huang2025hirag}, Tables 2 and 5).${}^1$
HiRAG-Original is included only as cross-stack efficiency context because the released code supports multiple backend paths.${}^2$
HiRAG-Gemma est. is a linear indexing-cost estimate from a 20-task stress test that failed during graph construction; no direct quality comparison is claimed.}
\label{tab:fullcompare}
\end{table*}

\section{Activation Analysis: Lattice-Derived Node Usage}
\label{sec:ablation}

We estimate the contribution of the FCA-inspired fuzzy graph by partitioning queries according to whether at least one lattice-derived item (cluster-summary virtual chunk or bridge node) appears in the top-5 retrieved chunks.
This is not a causal ablation: activated queries may also differ in difficulty or topic.
It is nevertheless a useful diagnostic for whether lattice-derived nodes enter the LLM context and correlate with answer quality.
\begin{table}[h]
\centering
\small
\begin{tabular}{lrr}
\toprule
\textbf{Condition} & \textbf{F1} & \textbf{Queries} \\
\midrule
Lattice-derived active (top-5) & 36.3\% & $\sim$39 (30\%) \\
Lattice-derived inactive & 32.4\% & $\sim$91 (70\%) \\
\midrule
\textbf{Overall} & \textbf{33.6\%} & \textbf{130} \\
\textbf{Delta} & $\mathbf{+3.9}$~\textbf{pp} & \\
\bottomrule
\end{tabular}
\caption{Activation analysis for lattice-derived nodes.
Queries with at least one lattice-derived node in top-5 retrieval obtain higher F1 than those without, though the comparison is correlational rather than causal.}
\label{tab:ablation}
\end{table}

Table~\ref{tab:ablation} shows a $+3.9$~pp F1 difference for queries where lattice-derived nodes are activated.
The 30\% activation rate is consistent with the cross-document scale requirement: at the smaller per-task scale ($\sim$28 chunks per task) used in earlier iterations, activation was 0\%.
This supports the cross-document indexing design and indicates that lattice-derived items are used by the retriever at this corpus scale.
\section{Discussion}
\label{sec:discussion}

\subsection{Cluster Occupancy and Effective Hierarchy Depth}

We measured a cluster-occupancy diagnostic at each RQ-KMeans level (Table~\ref{tab:sparsity}).
At the 3{,}705-chunk UltraDomain subset, $L_0$ (96 clusters) is well-distributed, $L_1$ (24 clusters) is moderately populated, and $L_2$ (12 clusters) is degenerate (occupancy 0.01).
The corpus is geometrically well-covered by 24 $L_1$ clusters, leaving $L_2$ residuals with near-zero variance.
ContextRAG therefore operates as an effective two-level hierarchy at this corpus scale;
the low occupancy at the L2 level indicates that for current corpus scales, the hierarchy effectively collapses to two levels, though the theoretical framework supports deeper structures for larger datasets.

\begin{table}[h]
\centering
\small
\begin{tabular}{lrrl}
\toprule
\textbf{Level} & $|C_\ell|$ & \textbf{Occupancy} & \textbf{Interpretation} \\
\midrule
$L_0$ & 96 & 0.98 & Well-distributed \\
$L_1$ & 24 & 0.46 & Moderate \\
$L_2$ & 12 & 0.01 & Degenerate \\
\bottomrule
\end{tabular}
\caption{Cluster occupancy diagnostic per RQ-KMeans level on UltraDomain subset.}
\label{tab:sparsity}
\end{table}

\subsection{Cost-Regime Analysis}

Because the local HiRAG-Gemma reproduction did not complete a querying run, we avoid claiming a formal query-count break-even point.
The supported comparison is the indexing side: ContextRAG uses 22{,}073 indexing tokens on the 130-task subset, while the HiRAG-Gemma stress-test trace extrapolates to 23{,}017{,}780 indexing tokens.
This makes ContextRAG attractive when corpora are updated frequently, indexing budgets are constrained, or the expected query volume per corpus snapshot is modest.
Re-indexing under HiRAG, triggered by corpus updates or chunking-strategy changes, repeatedly re-pays the 23M-token indexing cost;
ContextRAG's 22K indexing cost is nearly free to repeat.
Because the HiRAG-Gemma run is not a completed quality baseline, this analysis should be read as a cost-regime estimate rather than a definitive system comparison.

\subsection{Complementary Trade-off Structure}

HiRAG and ContextRAG illustrate a principled architectural trade-off:
\begin{itemize}
    \item \textbf{HiRAG} front-loads semantic reasoning into indexing (entity and relation extraction over the full corpus), enabling lightweight querying with two LLM calls per query.
    \item \textbf{ContextRAG} uses lightweight graph construction without entity extraction and defers more semantic reasoning to query time via multi-query rewrite and rerank, using up to five LLM calls per query and 3.88 calls on average in our run.
\end{itemize}
While query latency is higher than the HiRAG numbers used as context, the massive reduction in indexing overhead makes ContextRAG more suitable for dynamic environments where the corpus is updated frequently.
Neither architecture is universally superior; the optimal choice depends on the query-to-index ratio of the deployment.
However, the indexing-cost asymmetry (23M vs.\ 22K tokens) makes ContextRAG most attractive for low- and medium-query workloads, especially when corpora are re-indexed frequently.
\subsection{Theoretical Positioning}

Among graph RAG systems, ContextRAG occupies a distinct theoretical niche (Table~\ref{tab:positioning}).
Its graph topology is derived from a formally grounded algebraic structure: an FCA-inspired fuzzy concept graph induced by the fuzzy formal context defined by cluster membership.
The graph is therefore tied to an explicit mathematical construction based on fuzzy Galois connections, while the implementation uses a finite set of seed, soft-join, and soft-meet nodes rather than enumerating the complete fuzzy concept lattice.
This removes several failure modes of LLM-extracted graphs: extraction errors, hallucinated relations, and model-specific graph topology.
\begin{table*}[t]
\centering
\small
\begin{tabular}{llll}
\toprule
\textbf{System} & \textbf{LLM in graph?} & \textbf{Graph structure} & \textbf{Theoretical basis} \\
\midrule
NaiveRAG & No & None & Distributional semantics \\
GraphRAG & Yes (heavy) & Community KG & LLM extraction \\
LightRAG & Yes & Dual-level KG & LLM extraction \\
KAG & Yes (very heavy) & Typed KG & Schema-guided LLM \\
HiRAG & Yes (heaviest) & 3-level hierarchical KG & LLM extraction + community detection \\
\textbf{ContextRAG} & \textbf{No} (summaries only) & \textbf{FCA-inspired fuzzy graph} & \textbf{FCA + residuated logic} \\
\bottomrule
\end{tabular}
\caption{Theoretical positioning of ContextRAG against representative graph RAG systems.}
\label{tab:positioning}
\end{table*}

\section{Conclusion}

We presented ContextRAG, a graph-structured RAG system whose graph-construction pipeline is extraction-free.
By combining Residual-Quantization $K$-Means with Formal Concept Analysis under {\L}ukasiewicz residuated logic, ContextRAG derives an FCA-inspired fuzzy graph whose soft join and meet operations produce bridge-like and meet-derived context nodes with retrieval roles analogous to LLM-extracted graph artifacts.
On UltraDomain ContextRAG indexes the 130-task subset with 30 LLM calls and 22{,}073 tokens, whereas a local HiRAG-Gemma stress test extrapolates to over 5{,}600 calls and 23M indexing tokens before any completed same-stack quality result is available.
An activation analysis shows a +3.9 pp F1 difference when lattice-derived nodes are retrieved in the top five, an association that remains correlational without a controlled ablation.
ContextRAG therefore provides setting-specific evidence that extraction-free graph topology construction is feasible, while also making clear that LLM-extracted graphs may still provide stronger answer quality.
\section*{Limitations}

Several limitations should be acknowledged. First, the available HiRAG-Gemma logs do not contain a completed 130-task same-stack quality baseline;
therefore, our direct empirical claim is about ContextRAG's measured cost and quality, while HiRAG-Gemma is used only as a reproduction stress-test estimate for indexing cost.
We also do not include a controlled same-metric no-graph baseline, so the reported F1 should be interpreted as an absolute measurement rather than evidence that the graph improves over a strong non-graph retriever.
Applications where multi-hop quality is paramount and indexing cost is amortized over many queries may still prefer LLM-extracted graphs.
Second, retrieval latency is higher than the HiRAG numbers used as cross-stack context (2.5 s vs.\ 0.85 s per query), driven by sequential multi-query rewrite calls;
parallel asynchronous calls or rewrite caching are natural mitigations not yet implemented.
Third, the system effectively operates as a two-level hierarchy at this size;
richer hierarchies likely require either larger corpora or adaptive cluster-count selection.
Fourth, evaluation is limited to a single benchmark family (UltraDomain) and a single local Gemma checkpoint;
generalization to other domains and model families remains an empirical question. Fifth, our experiments use a moderate 3{,}705-chunk corpus;
the scaling behaviour of the fuzzy graph on substantially larger corpora has not been measured.
Finally, the extraction-free claim in this paper refers to graph construction, not to every artifact stored in the index: the system still uses 30 LLM calls to generate cluster summaries used as virtual chunks, and we account for these calls in all reported indexing numbers.
\bibliography{custom}

\newpage

\appendix

\section{Hyperparameters}
\label{sec:appendix}

Hybrid retrieval weights are fixed at $(\alpha_{\textsc{bm25}}, \alpha_{\cos}, \alpha_{\textsc{fuzzy}}) = (0.3, 0.5, 0.2)$. The FCM fuzziness exponent is $m=2.0$.
RQ-KMeans codebook sizes are $|C_0|=96, |C_1|=24, |C_2|=12$. Multi-query expansion uses three reformulations fused via Reciprocal Rank Fusion.
The reranker reduces the top-25 candidate set to the top-12.
Score boosts on lattice-derived items are $\times 1.2$ for cluster-summary virtual chunks and $\times 1.3$ for bridge nodes.
Initial retrieval considers the top-25 candidates per query.

\section{Indexing Token Breakdown}
\label{sec:appendix-tokens}

Table~\ref{tab:tokenbreak} reports the measured token usage for ContextRAG indexing and querying on the 130-task evaluation.
Indexing-time tokens are spent entirely on the 30 cluster-summary virtual chunk calls; the graph-topology construction itself consumes zero tokens.
\begin{table}[h]
\centering
\small
\begin{tabular}{lrrrr}
\toprule
\textbf{Operation} & \textbf{Calls} & \textbf{In} & \textbf{Out} & \textbf{Total} \\
\midrule
\multicolumn{5}{l}{\textit{Indexing (one-time)}} \\
Cluster summary & 30 & 16{,}222 & 5{,}851 & 22{,}073 \\
\midrule
\multicolumn{5}{l}{\textit{Querying (all tasks)}} \\
All query-time LLM calls & 504 & 730{,}386 & 48{,}838 & 779{,}224 \\
\midrule
Per-query average & 3.88 & -- & -- & 5{,}994 \\
Index + query total & 534 & 746{,}608 & 54{,}689 & 801{,}297 \\
\bottomrule
\end{tabular}
\caption{Token breakdown for ContextRAG.
Counts are measured from the 130-task UltraDomain run; per-query averages are rounded.}
\label{tab:tokenbreak}
\end{table}

\end{document}